\def\year{2018}\relax
\documentclass[letterpaper]{article} %DO NOT CHANGE THIS

\usepackage{aaai18}
\usepackage{times}
\usepackage{helvet}
\usepackage{courier}
\usepackage{url}
\usepackage{graphicx}
\usepackage{amsmath}
\usepackage{amssymb}
\usepackage{bm}
\usepackage{mediabb}

\def\etal{\emph{et al.}}

\newcommand{\argmin}{\mathop{\rm arg~min}\limits}

\begin{document}

\title{Weakly Supervised Collective Feature Learning from Curated Media}
\author{%
  Yusuke Mukuta $^{1,3}$, Akisato Kimura $^{1,2}$, David B. Adrian $^{1,4}$, Zoubin Ghahramani $^{2,5}$\\
  1. NTT Communication Science Laboratories, Japan.\\
  2. Department of Engineering, University of Cambridge, United Kingdom.\\
  3. The University of Tokyo, Japan.\quad
  4. Technical University of Munich, Germany.\quad
  5. Uber AI Labs, USA.\\
  {\tt\small mukuta@mi.t.u-tokyo.ac.jp, akisato@ieee.org, david.adrian@tum.de, zoubin@eng.cam.ac.uk}
}

\maketitle

\begin{abstract}
The current state-of-the-art in feature learning relies on the supervised learning of large-scale datasets consisting of target content items and their respective category labels.
However, constructing such large-scale fully-labeled datasets generally requires painstaking manual effort.
One possible solution to this problem is to employ community contributed text tags as weak labels, however, the concepts underlying a single text tag strongly depends on the users.
We instead present a new paradigm for learning discriminative features by making full use of the human curation process on social networking services (SNSs).
During the process of content curation, SNS users collect content items manually from various sources and group them by context, all for their own benefit.
Due to the nature of this process, we can assume that (1) content items in the same group share the same semantic concept and (2) groups sharing the same images might have related semantic concepts.
Through these insights, we can define human curated groups as weak labels from which our proposed framework can learn discriminative features as a representation in the space of semantic concepts the users intended when creating the groups.
We show that this feature learning can be formulated as a problem of link prediction for a bipartite graph whose nodes corresponds to content items and human curated groups, and propose a novel method for feature learning based on sparse coding or network fine-tuning.
\end{abstract}

%-------------------------------------------------------------------------
\section{Introduction}
\label{sec:intro}

Accuracy in image recognition and other related computer vision tasks strongly depends on the choice of image features.
Learning a discriminative feature representation is thus an essential step in handling these tasks.
Before the breakthrough brought by convolutional neural networks (CNNs), most of the approaches were based on hand-crafted feature representations.
However in the current state-of-the-art, this has changed to supervised feature learning from well-organized large-scale datasets consisting of millions of images and assigned labels fully annotated by experts \cite{He2015}.
However, constructing such large-scale datasets generally requires a huge amount of  manual effort and poses one of the biggest challenges in feature learning.
Instead, training features in a weakly supervised or completely unsupervised fashion allows us to utilize a very large amount of available data.
With the steadily increasing popularity of SNSs, large quantities of unlabeled and weakly labeled data are created every day.
Most existing research on image feature learning without relying on fully labeled image datasets utilizes community contributed text tags that can be easily collected through image search platforms \cite{Sukhbaatar2015} or content sharing platforms \cite{Joulin2016}.
However, the concepts underlying a single text tag depend strongly on the users who provide the text tag.
Different users may annotate the same text tag for different concepts, but we cannot distinguish those different concepts solely from the text tag, which may make it difficult to train discriminative features.

Instead, we present a new paradigm for learning visual features by making the full use of the human curation process on SNSs.
During the process of content curation, images are manually collected from many sources such as photo sharing services and other SNSs, grouped by concept, and provided with textual descriptions in the form of text tags or sentences \cite{Duh2012,Zhong2015}.
The concept of each image group is defined by the curating user, allowing for a wide range of very specific semantic concepts, such as "Green/Blue Ferrari" or "Gadgets for Lazy People", as shown in Figure \ref{fig:board}.
Previous research \cite{Zhong2015} focusing on Pinterest, which is one of the most popular content curation platform for images, has demonstrated that most of the images collected in the same curated group can be assumed to share the same semantic concept.
This readily implies that we can treat each human-curated group as a pseudo category that shares a consistent semantic concept compared with noisy community-contributed text tags.
However, employing curated groups as weak labels still presents several technical problems. First, we have to handle a large number of curated groups, as when using community-contributed text tags. Second, the number of images in a curated group is much smaller than what is required to fully capture the characteristics of the curated group.

\begin{figure*}
  \begin{center}
    \includegraphics[width=0.975\hsize]{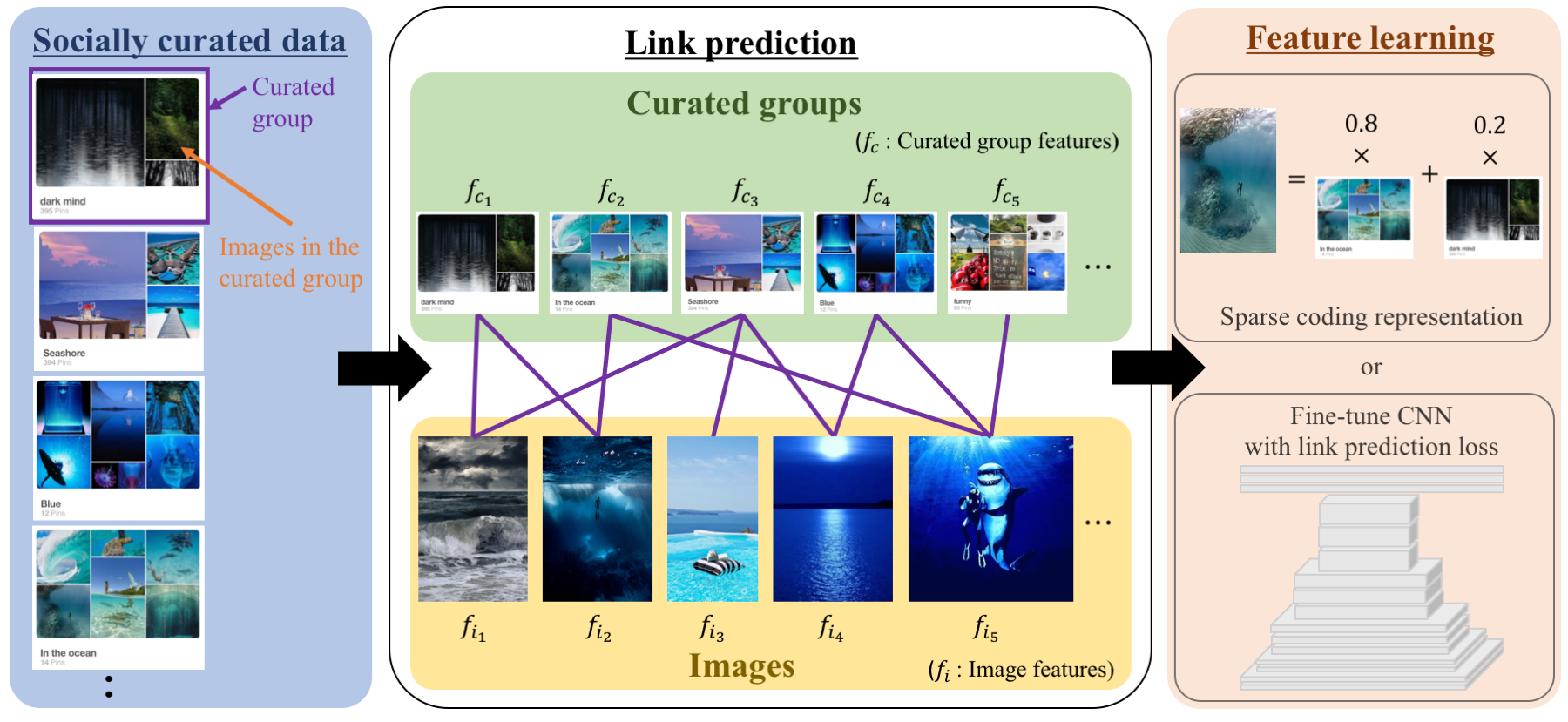}
  \end{center}
  \caption{%
    Overview of the proposed collective feature learning from curated media.
    Images in a curated group share the same concept and curated groups sharing the same image have a similar concept.
    On the basis of these insights, we can regard curated groups as weak labels for content, and formulate the problem of estimating weak labels as a link prediction problem for a bipartite graph, where one node group corresponds to images and the other is for curated groups.
    The models for link prediction and a new feature representation can be iteratively optimized.
  }
  \label{fig:system}
\end{figure*}

In this paper, we propose a novel framework for weakly supervised image feature learning that fully exploits the properties underlying human curated content.
Figure \ref{fig:system} shows an overview of the proposed method.
We introduce another property of socially curated data that results from the fact that curated groups sharing several images are expected to have similar concepts.
Namely, to capture the nature of a curated group we can utilize not only images in the group but also those in neighboring curated groups.
We can see that the problem of predicting to which curated groups a new image belongs can be regarded as multi-task learning, where each task corresponds to the classification of images into a curated group.
In addition, we note that the task of classifying images into curated groups can also be seen as classification using only positive and unlabeled data (PU classification) \cite{dePlessis2015}, since we can find many curated groups with the same concept but sharing no images with each other, due to the locality of SNSs.
Link prediction techniques have been proved to be effective for this PU classification \cite{Chang2016} and can be applied to large-scale graphs \cite{Menon2011}.
Thus, we formulate our feature learning as the problem of predicting links on a bipartite graph, where one node group corresponds to images and the other corresponds to curated groups.
We then propose a novel method for link prediction that exploits observed features on nodes as well as link structures.
In a similar manner to learning visual attributes \cite{Farhadi2009}, we can apply scores of linking curated groups to a given image as a new feature of the image.
However, we propose a more sophisticated method for learning a new feature representation, which combines a model for link prediction and convolutional neural networks for feature extraction.
Although we concentrate on the use of a specific SNS platform for still images mainly for presentation clarity, our method can be easily applied to other platforms and modalities, as long as each user collects contents into groups sharing the same semantic information.
For example, users in music platforms such as Spotify, Amazon Music and iTunes can create playlists that are such curated groups.
The underlying graph can then be used, e.g., to improve existing approaches for music feature learning with deep CNNs (see e.g. \cite{Lee2009}).
We also note that our fine-tuning approach can be directly applied to other types of graph structures representing relationships among (pseudo) labels, such social networks or knowledge graphs.

%-------------------------------------------------------------------------
\section{Related work}
\label{sec:related}

%%--------
\subsection{Feature Learning}
\label{sec:related:learning}
\indent

Image feature learning is a well-studied problem in the computer vision community.
CNN architectures VGGNet \cite{Simonyan2014} are capable of capturing object-level semantics, but require very large labeled datasets for proper training.
This background encourages computer vision researchers to develop weakly supervised feature learning.
This type of feature learning can be categorized into two approaches, namely (1) collecting or generating weakly supervised information and (2) transferring knowledge obtained from other modalities such as natural languages \cite{Socher2013,Silberer2014,Hoffman2016,Changpinyo2016}.
The work presented in this paper focuses on the first approach, since it can work well for uni-modal situations and can be easily integrated into the second approach.
The first approach often employs web images since a huge amount of meta-data and side information is accessible.
Image search engines are one of the most common and the easiest interfaces with regards to collecting web images with weak text labels \cite{Sukhbaatar2015}, where words or phrases for querying images can be used as weak labels.
However in this strategy, rich meta-data and other side information underlying query texts and images are not accessible, and thus ad-hoc and complicated data cleansing would need to be developed \cite{Yashima2016}.
Meanwhile, the emergence of SNSs has enabled us to collect rich meta data and side information easily as well as images and associated text words, phrases and sentences.
Recent research \cite{Joulin2016} has demonstrated that text information associated with images is useful for training image features with deep learning.
However, the text tags or phrases provided by SNS users are still noisy in the same way as those obtained via image search engines, which encourages researchers to make the full use of the meta-data and side information available on SNSs.
\cite{Yuan2013} proposed a relational generative deep learning model integrating multi-modal features and their relationships.
\cite{Geng2015} tried to learn feature representations for both users and images on SNS by transforming the heterogeneous user-image networks into homogeneous low-dimensional representations.
\cite{Fang2015} and \cite{Zhang2016} employed a collaborative filtering approach to define latent classes based on user behavior data, and use these classes as labels to train a supervised classifier.

Our proposed framework for collective feature learning focuses on the use of content groups curated by individual users, which are different and less noisy data resources for weakly supervised feature learning compared with noisy community contributed tags or user behavior.
Karayev \etal \cite{Karayev2014} employed Flickr Groups that are community curated collections of visual concepts to learn and recognize the visual styles of images or paintings, such as Baroque, Cubism and Impressionism.
Such community-curated collections are highly useful for predicting well-known or well-defined class categories, since almost all the curators share the same concept for each category.
Instead, we focus on the use of user-curated groups that are expected to be more focused and consistent in terms of concepts since this enables us to apply a broad range of tasks.

Node2vec \cite{Grover2016} is closely related to our method since it provides a way of learning feature representations for graph nodes.
However, this only employs graph structures such as connections between nodes and edge weights for representation learning, and the integration of observations on nodes is not straightforward.
The use of image features as observations on nodes is crucial in terms of our problem setting for collective feature learning.

%%--------
\subsection{Link Prediction}
\label{sec:related:link}
\indent

The task of link prediction is to predict the presence or absence of unobserved edges between nodes in a graph from already observed edges, and it has been used to analyze the characteristics of social networks and to recommend new items for customers.
Meanwhile, this paper provides a novel application of link prediction that incorporates it into the learning of image features from social activities on SNSs.

Link prediction can be categorized into unsupervised and supervised approaches.
Unsupervised methods compute a score of links between a node pair based on a pre-defined score function and predict the presence of edges based on the computed score such as common neighbors \cite{Kossinets2006}, Katz index \cite{Katz1953} and SimRank \cite{Jeh2002}.
Although unsupervised methods do not require any pre-processing or training processes, their performance is unsatisfactory if we cannot choose an appropriate score function that suits a given network.
Supervised methods utilize already observed edges as training labels and learn a parameterized score function so that the learned score function predicts the observed edges well.
Several types of score functions have already been proposed.
\cite{Wang2007} exploited a weighted sum of node attributes as a score function.
\cite{Kashima2009} proposed a graph kernel between node pairs.
\cite{Scripps2008} assumed that each node has an observed feature, and proposed a model that reflects the similarity between observed features on nodes.
\cite{Menon2011} introduced the idea of latent features on nodes, and proposed a matrix factorization method for link prediction.
Our proposed method extends Menon and Elkan's method so as to effectively extract image features.

%-------------------------------------------------------------------------
\section{Content curation}
\label{sec:curation}

In this section, we describe the process of content curation and dataset collection obtained with a content curation platform and its unique properties.

%--------
\subsection{Characteristics}
\label{sec:curation:def}
\indent

\begin{figure}
  \begin{center}
    \includegraphics[width=\hsize]{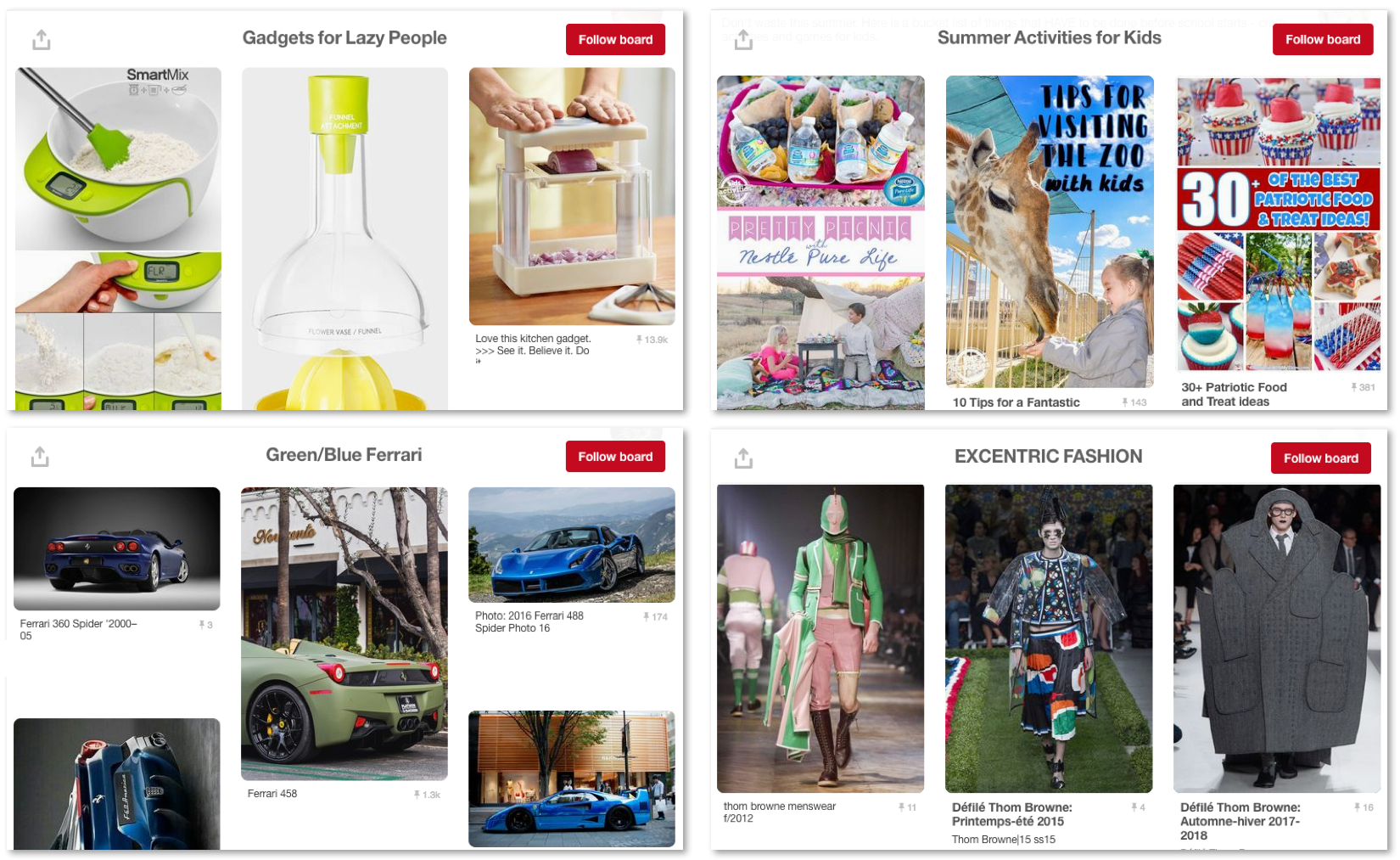}
  \end{center}
  \caption{Several examples of curated image groups, extracted from Pinterest.}
  \label{fig:board}
\end{figure}

Many SNSs have been created to connect people and allow them to exchange general information and to share items of personal interest.
\textit{Content curation platforms} are a special sub-category of SNS.
Content curation can be defined as a spontaneous human process that consists of remixing social media content for the purpose of further consumption.
What characterizes content curation is the manual efforts made by social media users involved in organizing social media content.
Users collect various types of media content items, re-group them according to their own preferences, provide additional text tags or descriptions, and publish them on SNSs, all for their own benefit.
A curated group of content items is not only simply aggregated, but also extended with additional information and placed into a new semantic context, solely decided by the curating user.
A published curated group allows other users to reuse a part of the group to organize another curated group.
Fig \ref{fig:board} shows several examples of curated image groups.

Curated groups have several favorable properties for our purpose in this research, namely weakly supervised feature learning.
First, the process of content curation can be regarded as a human computation process of content classification, and thus the content in the same curated group shares a similar concept or meaning \cite{Zhong2015}.
This readily implies the potential of socially curated groups as weak, but promising labels.
Second, content can go viral on social media as well as content curation platforms, and as a result that content is distributed from group to group, which constitutes a content-centric graph structure.
The above two properties imply that two curated groups sharing a lot of content often share a similar concept \cite{Kimura2013}.
Our proposed method makes full use of these insights for training image features without the need for the painful effort of building labeled image datasets.

%--------
\subsection{Dataset}
\label{sec:curation:data}
\indent

\def\a{\bm{a}}
\def\b{\bm{b}}
\def\s{\bm{s}}
\def\x{\bm{x}}
\def\y{\bm{y}}
\def\z{\bm{z}}
\def\A{\bm{A}}
\def\B{\bm{B}}
\def\V{\bm{V}}
\def\W{\bm{W}}
\def\X{\bm{X}}
\def\Y{\bm{Y}}
\def\Z{\bm{Z}}
\def\bR{\mathbb R}
\def\cC{\mathcal C}
\def\cT{\mathcal T}
\def\cU{\mathcal U}
\def\cL{\mathcal L}

For learning our model, we crawled content curation data from Pinterest, a popular content platforms specializing in image curation \footnote{We will publish the dataset at \url{http://www.kecl.ntt.co.jp/people/kimura.akisato/socialweb4.html}.
}.
In Pinterest, a curated group is called a \emph{board} and each element assigned to a curated group is called a \emph{pin}.
Each board has a text title, and each pin contains a single image and a description provided by the curating user.
Without loss of generality, we consider each pin as a feature vector $\x\in\bR^{D_I}$.
This feature vector could be extracted from the image itself, the textual description, or any other feature representations thereof, but in the following $\x$ is a feature vector extracted from an image.
We also consider each board as a feature vector $\y\in\bR^{D_C}$, and in the following $\y$ is a feature vector extracted from the name of the board.

\begin{figure}[t]
  \centering
  \includegraphics[width=0.9\hsize]{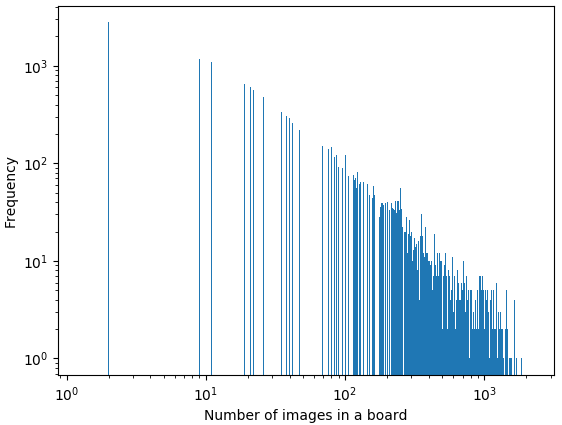}
  \includegraphics[width=0.9\hsize]{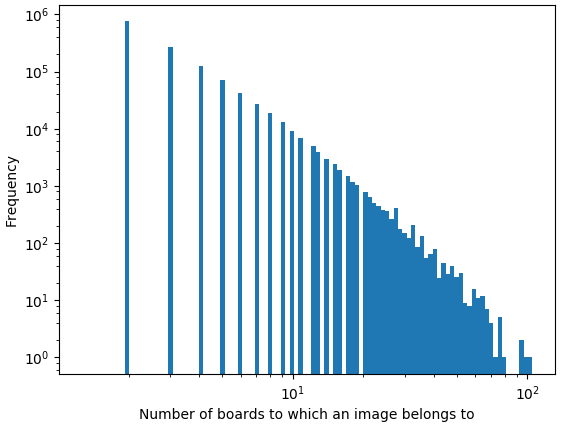}
  \caption{%
    (Top)    Distribution of number of images in individual boards
    (Bottom) Distribution of number of boards to which individual images belong
  }
  \label{fig:degs}
\end{figure}

Pinterest provides 36 default categories that cover a broad range of topics such as architecture, entertainment, geek, outdoors, quotes and weddings.
We started the data collection by taking the most recent pins in each of the 36 pre-defined categories.
For example, recent pins for the “popular” category can be found at ttps://www.pinterest.com/categories/popular. %\url{https://www.pinterest.com/categories/popular}.
Branching out from the set of users who posted these recent pins, their friends, namely followers and followees, were also collected as new seeds, which results in a set of $6.7$K users.
Next, we collected all the boards curated by each individual user.
We skiped all boards that are either under-populated, e.g., they have fewer pins, or if they are much larger than the average, since it is questionable how representative they can be for feature learning.
Then, we collected all the pins in each board.
Finally, we omitted image nodes with degrees of less than $2$ (meaning that those images belong to only $1$ curated group) and curated group nodes with degrees of less than $1$ (meaning that no images belongs to those curated groups), which are thought to be of no use for the task.
This dataset constitutes a total of $65$K boards and $8.2$M pins, meaning that $1.5$K pins per user on average.
It seems that too many images have been pinned, however, this is not surprising due to the nature of this collection scheme.
Namely, around $82$\% of the collected pins are \emph{repins}, e.g., carbon copies of other pins, which leave around $1.5$M unique pins.
Consequently, many images are assigned to two or more boards, each possibly within a different semantic context and offering a unique textual description.

Figure \ref{fig:degs} shows a distribution of numbers of images in individual boards and a distribution of numbers of boards to which individual images belong.
You can see that both of the distributions obey a power law.

%-------------------------------------------------------------------------
\section{Proposed method}
\label{sec:proposed}

In this section, we propose a novel method that makes full use of the potential of human curated data.

%--------
\subsection{Framework}
\label{sec:proposed:framework}
\indent

We provide an overview of our method in Figure \ref{fig:system}.
Training data $\cT$ contains $N_I$ images, $N_C$ curated groups, and a bipartite graph representing which image belongs to which curated groups.
Thus, we can represent the training data $\cT$ as a triplet of an image feature matrix $\X=(\x_i\in\bR^{D_I})_{i=1}^{N_I}$, a curated group feature matrix $\Y=(\y_c\in\bR^{D_C})_{c=1}^{N_C}$, and an adjacency matrix $\A\in\{0,1\}^{N_C\times N_I}$.
We used activations from the 7th fully-connected layers of the pre-trained VGG16 network \cite{Simonyan2014} for image feature extraction, and extract a distributed representation of the names of boards and users with the Word2Vec model \cite{Mikolov2013} for curated group features.
If we can predict a score vector $\hat{\a}_i\in\bR^{N_C}$ that represents the possibility of an image $i$ belonging to each of curated groups, we can use this score vector as a basis for novel features of the image that consider a broad range of concepts contained in the curated groups.

Our proposed method formulates the problem of predicting a score vector $\hat{\a}_i$ for an image $i$ with the framework of link prediction, where we train a model for link prediction with a given set of training data $\cT=(\A,\X,\Y)$.
Standard link prediction models try to estimate the presence of links between existing nodes whose links are missing.
Meanwhile, our setting needs to estimate the presence of links from a new node (i.e. a new image) to existing ones (i.e. curated groups), for the sake of collective feature learning, and for this reason it requires observed features for link prediction.
We propose a new link prediction model for this purpose that integrates the observed features of nodes as well as the structure of existing links.
Our method can iteratively optimize the models for link prediction and image feature extraction, which we will discuss in the following sections.

%--------
\subsection{Model for link prediction}
\label{sec:proposed:link}
\indent

In this section, we describe our proposed model for link prediction as a basis for collective feature learning.

Our model is based on a previous model proposed by \cite{Menon2011} that employs the observed and latent features of nodes and scales in terms of network sizes.
More formally, a score $\hat{a}_{i,c}$ of a link between an image node $i$ and a curated group node $c$ is modeled as
\begin{eqnarray}
  \hat{a}_{i,c} &=& \z_{Ii}^{\top}\W\z_{Cc} + \x_i^{\top}\V\y_c + b_{Ii} + b_{Cc},
  \label{eq:linkscore:old}
\end{eqnarray}
where $\z_{Ii}\in\bR^{D_I}$ and $\z_{Cc}\in\bR^{D_C}$ are latent features of the image node $i$ and the curated group node $c$, respectively, $b_{Ii}\in\bR$ and $b_{Cc}\in\bR$ are bias terms of the image node $i$ and the curated group node $c$, respectively, and $\W\in\bR^{D_I\times D_C}$ and $\V\in\bR^{D_X\times D_Y}$ are weight parameters.
This model assumes that there tends to be a link between nodes with a larger score, where the first term reveals the network structure by factorizing the ground-truth link matrix $\A$, and the second term reflects the feature similarity between nodes.
For a given set of training data $\cT$, all the model parameters $\W$, $\V$, $\Z_I=(\z_{Ii})_{i=1}^{N_I}$, $\Z_C=(\z_{Cc})_{c=1}^{N_C}$, $\b_I=(b_{Ii})_{i=1}^{N_I}$ and $\b_C=(b_{Ci})_{c=1}^{N_C}$ are optimized to minimize the following expected rank loss:
\begin{eqnarray}
  \sum_i \sum_{(c,c'):A_{i,c}=1,A_{i,c'}=0}
    \hspace{-8mm}l(\hat{a}_{i,c}-\hat{a}_{i,c'}) + \Omega(\W, \V, \Z_I, \Z_C),
  \label{eq:linkloss}
\end{eqnarray}
where $l(\cdot)$ is a loss function and $\Omega(\cdot)$ is a regularization term.
Although any types of loss functions and regularizations can be applied, in this paper we exploit a hinge loss for the loss function and $\ell_2$-norm regularization of all the model parameters except the bias terms.
Parameter updates can be easily derived by stochastic gradient descent.
For the computational efficiency, we update parameters for randomly select a pair $(c,c')$ of curated groups for each image, instead of computing the rank loss for all the possible combinations of curated groups.
In this model, a network structure $\A$ is encoded into the model parameter $\W$ and the latent variables $\Z_I$ and $\Z_C$, and the contributions of the observed features $\X$ and $\Y$ are concentrated onto another model parameter $\V$.
Therefore, this model is unsuitable for our purpose of collective feature learning, since the latent feature $\z_{\rm{new}}$ of a new image is unknown, and thus the network structure in the training data is totally ignored without the latent feature term.

Our proposed model extends the model proposed by Menon and Elkan to collective feature learning.
More specifically, a new score function of our proposed model is
\begin{eqnarray}
  \hat{a}_{i,c}
  &=& \begin{pmatrix}\x_i^{\top} & \z_{Ii}^{\top}\end{pmatrix}
      \begin{pmatrix}\W_X \\ \W_Z\end{pmatrix}
      \begin{pmatrix}\y_c \\ \z_{Cc}\end{pmatrix} + b_{Ii} + b_{Cc},
\end{eqnarray}
where $\W_X\in\bR^{D_X\times(D_Y+D_C)}$ and $\W_Z\in\bR^{D_I\times(D_Y+D_C)}$ are model parameters.
The difference between the previous and proposed models appears marginal, however, they are totally different in terms of their underlying concepts.
Our proposed model enables us to incorporate the network structure into link prediction even if there is no latent feature for the new image.
Although we developed the proposed model for collective feature learning, it improves the performance of link prediction by considering the correlations between observed and latent features.
The experimental results for link prediction will be presented later.
Through the experiments for link prediction, we found that the contribution of image latent features $\Z_I$ to the performance of link prediction was minor.
We also note that in our setting of collective feature learning the latent feature of a new image is unknown, and thus it cannot be used for the test stage.
Therefore, we use the following simplified model, which omits image latent features.
\begin{eqnarray}
  \hat{a}_{i,c}
  &=& \x_i^{\top} \W_X \begin{pmatrix}\y_c \\ \z_{Cc}\end{pmatrix} + b_{Ii} + b_{Cc}.
  \label{eq:linkscore:new}
\end{eqnarray}
This formulation can also be viewed as a multi-task linear classifier composed of a shared parameter $\W_X$, a class-dependent term $\z_{Cc}$, and a class-wise observed feature $\y_c$.

%--------
\subsection{Feature learning}
\label{sec:proposed:learning}
\indent

Once we have obtained a model for link prediction, we are ready to train a new feature representation for images.
One possible but naive way is simply to use a score vector $\hat{\a}_{\rm{new}}\in\bR^{N_C}$ as a new feature representation for a new image.
However, we propose two other approaches that are more sophisticated and effective.

The first approach employs sparse coding to represent an observed image feature $\x$ as a sparsely weighted sum of curated group features.
The proposed score function Eq. \eqref{eq:linkscore:new} for link prediction implies that an image feature $\W_X\x_i$ weighted by a model parameter $\W_X$ should be similar to a curated group feature $(\y_c^{\top},\z_{Cc}^{\top})^{\top}$ if the image $i$ belongs to the curated group $c$.
We also expect a single image to belong to only a few curated groups, since the image has only a few concepts and each curated group is expected to be focused in terms of concepts.
These two insights readily imply that a given image feature $\x$ can be represented as a mixture of a small number of curated group features, and we employ the coefficient $\bm{\alpha}(\x)\in\bR^{N_C}$ of the following sparse coding as a new feature representation of the image $\x$:
\begin{equation}
  \bm{\alpha}(\x)
  = \argmin_{\bm{\alpha}\in\bR^{N_C}} \left\{
    \left\|\W_X\x - \begin{pmatrix}\Y \\ \Z_C \end{pmatrix}
    \bm{\alpha} \right\|_2^2 + \lambda\|\bm{\alpha}\|_1 \right\}
\end{equation}
where $\lambda$ is a regularization parameter.
We adopt a simple thresholding method for sparse coding, and its threshold is determined by 10-fold cross-validation.

As described above, the first approach utilized a fixed image feature extractor, and it can be regarded as a kind of feature transformation.
This means that it can remove redundant aspects of raw image features but it cannot boost the intrinsic representation power.
On the other hand, the second approach fine-tunes the VGG16 network with the help of link prediction, which can potentially boost the representation power in principle.
With this approach, we expect a new feature to clearly distinguish a curated group from others with different concepts.
On the basis of the above discussion, we employ the rank loss shown in Eq. \eqref{eq:linkloss} as a loss function for optimizing image features, where the network structure remains the same as VGG16.
Since the joint optimization of the models for image feature extraction and link prediction is not straightforward, we follow the alternative optimization approaches that are widely used for training deep composite models.
The link prediction model is first trained with a fixed image feature model, namely a pre-trained VGG16 network.
The rank loss Eq. \eqref{eq:linkloss} is then back-propagated to the VGG16 model for fine-tuning of the overall VGG16 network.
The link prediction model can be again trained for the fine-tuned image feature model if required.
Finally, we use the output of the activations of the 7th fully-connected layer FC7 (4,096 dimensions) in the fine-tuned VGG16 model as a new feature representation.

The above two approaches provide aspects of feature representations that are different from VGG image features trained with the object recognition task.
Thus, our new feature representations are complementary with the pre-trained VGG16 features, and combining two different feature representations will boost the performance.

%-------------------------------------------------------------------------
\section{Experiments}
\label{sec:exp}

In this section, we evaluate the effectiveness of the proposed method through several experiments.

\begin{table*}[t]
  \caption{Performance of link prediction measured by mean AUC}
  \label{table:exp_link}
  \begin{center}
    \begin{tabular}{|l|r|r|r|r|r|}\hline
      Size                                 & Baseline & No latent & Latent image & Latent group & Latent both \\ \hline
      Small ($2.4$K groups, $850$K images) & 0.895    & 0.898     & 0.896        & 0.928        & 0.929       \\ \hline
      Large ($65$K groups, $1.5$M images)  & 0.8442   & 0.8441    & 0.8460       & 0.8482       & 0.8477      \\ \hline
    \end{tabular}
  \end{center}
\end{table*}

%\begin{figure}[t]
%  \begin{center}
%    \includegraphics[width=0.985\hsize]{fig/exp_link.png}
%  \end{center}
%  \caption{Performance of link prediction measured by mean AUC}
%  \label{fig:exp_link}
%\end{figure}

\begin{table*}[t]
  \caption{Classification performances on test data}
  \label{table:classify}
  \begin{center}
    \begin{tabular}{|c|c|c|c|c|c||c|c|c|c|}\hline
      Dataset    & Metric & VGG   & FT-GRP & FT-WORD & LINK  & PROP-SC & PROP-FT & VGG+SC & VGG+FT \\ \hline
	  Pinterest  & MAP    & 0.430 & 0.044  & 0.043   & 0.382 & 0.409   & 0.404   & 0.445  & {\bf 0.477}  \\
      UECFOOD100 & ACC    & 0.511 & 0.051  & 0.051   & 0.065 & 0.454   & 0.453   & 0.514  & {\bf 0.536}  \\
      UECFOOD256 & ACC    & 0.468 & 0.023  & 0.023   & 0.028 & 0.376   & 0.398   & 0.473  & {\bf 0.502}  \\
      Apparel    & ACC    & 0.576 & 0.146  & 0.144   & 0.319 & 0.488   & 0.470   & 0.566  & {\bf 0.586}  \\
	  Hipster    & ACC    & 0.592 & 0.363  & 0.244   & 0.345 & 0.552   & 0.615   & 0.599  & {\bf 0.642}  \\
	  Instagram  & MAP    & 0.910 & 0.783  & 0.785   & 0.868 & 0.912   & 0.896   & 0.916  & {\bf 0.919}  \\ \hline
    \end{tabular}
  \end{center}
\end{table*}

%--------
\subsection{Link prediction performance}
\label{sec:exp:link}
\indent

We first evaluate the performance of the proposed link prediction model.
We employed  the dataset described in Section \ref{sec:curation:data} and constructed a bipartite graph with $65$K group nodes and $1.5$M image nodes.
We also built a smaller graph with $2.4$K group nodes and $850$K image nodes for evaluating the effect of network sizes.
We used the output of the FC7 layer of the VGG16 network as image features with $4,096$ dimensions.
To extract features from curated groups, we first split the user names and titles of the curated groups into words, and then used the mean of the word vectors extracted from the Word2Vec model \cite{Mikolov2013}.
We removed out-of-vocabulary words when computing the mean vector, and if no words existed for computing the mean we simply remove the curated group nodes.
The word vector model was trained with GoogleNews and the number of dimensions was set to 300 (default).
To separate the dataset into training and test data, we first selected $10$\% (or $1$ if the node had less than $10$ edges) of all the edges for each image node as test data, and used the rest for training.
The baseline for comparison was the model proposed by \cite{Menon2011}, and we have examined the performance of our proposed method with and without using latent features.
We adopt the mean of area under the ROC curve (AUC) for each image as a measure for evaluation.

Table \ref{table:exp_link} shows the results.
The result indicates that our proposed method outperformed the previous method and the introduction of latent variables for curated groups provided the greatest improvement to link prediction performance.
On the other hand, image latent variables did not produce significant improvements in the performances.
This may be because image features extracted from VGG were sufficiently informative for predicting links, and we did not need to learn any additional latent features.
On the basis of this result, we used our model Eq. \eqref{eq:linkscore:new} only with only the latent features of the curated groups in the following experiments.

Though the performance gain of our proposed method on the large network looks marginal especially for the large graph, the test data for the large graph contains $9.75$G possible links ($= 65$K curated groups $\times 1.5$M images $\times 10$\% kept for test), and thus $0.4$\% improvements affect approximately $39$M possible links that are much larger than the number of images.
Therefore, this performance gain is significant and has a great impact on both link prediction and feature learning.
We also note that links that can be easily estimated only with image features tend to increase as the network size increases.

%%--------
\subsection{Experiment using learned feature}
\label{sec:exp:pred}
\indent

Next, we tested our proposed method on various image classification tasks.
%
%To validate the effectiveness and generalization performance of the proposed method, we used various datasets from several different domains, such as food classification (Food101 \cite{Bossard2014}, UEC-FOOD100 \cite{Matsuda2012}, UEC-FOOD256 \cite{Kawano2014c}), fashion classification (Hipster Wars \cite{Kiapour2014}, Apparel \cite{Bossard2012}), scene classification (SUN397 \cite{Xiao2010}) and image sentiment analysis (Instagram \cite{Katsurai2016}).
To validate the effectiveness and generalization performance of the proposed method, we used various datasets from several different domains, such as food classification (UEC-FOOD100 \cite{Matsuda2012}, UEC-FOOD256 \cite{Kawano2014c}), fashion classification (Hipster Wars \cite{Kiapour2014}, Apparel \cite{Bossard2012}) and image sentiment analysis (Instagram \cite{Katsurai2016}).
In addition, we collected a new dataset from Pinterest for evaluation, which was collected independently from the one used for training the whole model.
Pinterest has 36 default categories that were accessible from the top page.
We excluded 4 categories that do not relate to the content, namely ``Popular'', ``Everything'', ``Videos'' and ``Quotes'', and used the remaining 32 categories as a class.
We collected images that are pinned to the boards that belong to those categories and constructed a dataset with $63$K images and $32$ classes.
We followed the separation of the training and test data given by the dataset distributors for almost all the datasets excluding Instagram and Pinterest, and we randomly separated the training and test data for the $2$ remaining datasets.
More specifically, we used $20$K training samples and $20$K test samples for the Instagram dataset and $500$ training samples per each class, namely $10$K images in total, for the Pinterest dataset.

We compared image features extracted from
(1) VGG: the VGG16 model as a baseline,
(2) FINE-GRP: a fine-tuned VGG16 model in which we consider curated group assignments as class labels (for simulating the proposed method without link prediction),
(3) FINE-WORD: a fine-tuned VGG model in which we consider words contained in a pin description as class labels (for simulating the previous research \cite{Joulin2016}),
(4) LINK: a collection of link prediction scores $\hat{a}_{i,c}$ as a new image feature for image $i$ with $N_C$ dimensions (= the number of curated groups),
(5) PROP-SC: our proposed method with the sparse coding approach and
(6) PROP-FINE: our proposed method with the fine-tuning approach.
To train the FINE-WORD model, we excluded stop words with NLTK and selected the top $10$K words as a set of class labels.
We also tried a late fusion of our proposed feature representations and the pre-trained VGG feature.
As a metric for our evaluations, we used the mean average precision for the Pinterest (multi-label prediction) and Instagram (binary classification) datasets and the classification accuracy for the remaining datasets (multi-class categorization).

Table \ref{table:classify} summarizes the results.
The results indicate that FT-WORD features trained with words in pin descriptions as class labels, like the previous method \cite{Joulin2016}, did not produce meaningful features, since pin descriptions brought by SNS users are often noisy, and feature learning based on such noisy supervised information often fails.
The experimental results also indicates that FT-GRP features based on the proposed method without link prediction also failed, since many curated groups contain only a small number of images, as shown in Figure \ref{fig:degs}, and thus the model easily overfit the training data.
On the other hand, the classification performance of our method using sparse coding was comparable to that of the original VGG16, and the late fusion with VGG16 produced better performances.
This implies that our proposed sparse coding feature conveys discriminative information that is not contained in the original VGG feature.
We can also see from the table that our proposed feature based on the fine-tuning approach performed comparable to VGG16 even without any strong supervised information for feature learning, and the late fusion with VGG16 performed the best for all the datasets.
This result justifies our statement that our features were learned to represent a broad range of user interests and social trends that might be difficult to explicitly describe by texts, and thus our features complement the original VGG features tuned for object recognition.

%-------------------------------------------------------------------------
\section{Conclusion}
\label{sec:conclude}

We have proposed a novel method for automatically learning discriminative level feature representations with the help of a massive amount of human-curated content on SNSs.
We have also introduced a way of combining multiple modalities in a weakly supervised learning setting.
To exploit the property of social curation data whereby two images in the same group share the same semantic concept and groups sharing the same image might have related semantic concepts, we proposed a novel framework that regards the relationship between images and concepts as a graph and learns image features via a link prediction problem.
We proposed a novel link prediction method that uses the weighted product of observed node attributes and learned latent features.
We also proposed a method for extracting a novel image feature from a learned network model.
We crawled image datasets from Pinterest for use as source curation data.
We trained our link prediction model from the dataset, applied learned image features to various benchmark datasets, and showed the effectiveness of the learned feature.
These results show that our novel link prediction-based framework is promising.

Since our prediction method is simple, more sophisticated methods may provide more discriminative features.
Also, we used only Pinterest as source curation data in our research, but we think that most social curation media have similar structures thus making our method applicable.
Analyzing the relation between the source media and the learned image feature will constitute our future work.

%-------------------------------------------------------------------------
% references section

\bibliographystyle{aaai}
\bibliography{aaai18mukuta}

% that's all folks
\end{document}